\documentclass[letterpaper, 10 pt, conference]{ieeeconf}  
\usepackage{multirow}
\usepackage{graphicx} 
\usepackage{stackengine} 
\usepackage{xcolor}
\usepackage{subfig}
\usepackage{amssymb}
\usepackage{amsmath} 
\usepackage{algorithm}
\usepackage{algorithmic}
\usepackage{url}

\IEEEoverridecommandlockouts                              

\title{\LARGE \bf
ROA-BEV: 2D Region-Oriented Attention for BEV-based 3D Object Detection
}

\author{Jiwei Chen$^{1}$, Yubao Sun$^{2}$, Laiyan Ding$^{1}$ and Rui Huang$^{1*}$
\thanks{$^{*}$Corresponding author}
\thanks{$^{1}$Jiwei Chen, Laiyan Ding and Rui Huang are with School of Science and Engineering, The Chinese University of Hong Kong, Shenzhen 518172, China (e-mail: \{jiweichen2, laiyanding, ruihuang\}@link.cuhk.edu.cn).}
\thanks{$^{2}$Yubao Sun is with the Engineering Research Center of Digital Forensics, Ministry of Education, Nanjing University of Information Science and Technology, Nanjing 210044, China (e-mail: sunyb@nuist.edu.cn)}
}

\begin{document}

\maketitle
\thispagestyle{empty}
\pagestyle{empty}

\begin{abstract}
Vision-based \textbf{Bird's-Eye-View (BEV)} 3D object detection has recently become popular in autonomous driving. However, objects with a high similarity to the background from a camera perspective cannot be detected well by existing methods. In this paper, we propose a BEV-based 3D Object Detection Network with \textbf{2D Region-Oriented Attention (ROA-BEV)}, which enables the backbone to focus more on feature learning of the regions where objects exist. Moreover, our method further enhances the information feature learning ability of ROA through multi-scale structures. Each block of ROA utilizes a large kernel to ensure that the receptive field is large enough to catch information about large objects. Experiments on nuScenes show that ROA-BEV improves the performance based on BEVDepth. The source codes of this work will be available at \url{https://github.com/DFLyan/ROA-BEV}. 

\end{abstract}

\section{Introduction}
The surround-view camera system is currently  one of the most popular sensor systems in vision-based autonomous driving solutions. \textbf{Bird's-Eye-View (BEV)} is a typically surround-view camera sensor system. This system generates a top-down view of the surrounding environment of the vehicle through the fusion of multiple cameras. Through the sensed images, autonomous vehicles can perform 3D object detection tasks to understand external scenes. It can detect the 3D attributes of objects, such as coordinates and size. A typical BEV-based 3D object detection model includes an image backbone, a \textbf{View Transformation Module (VTM)}, and a 3D object detection head. Specifically, the backbone contains a feature extraction module, such as ResNet \cite{resnet}, and a feature fusion module, such as \textbf{Feature Pyramid Networks (FPN)} \cite{fpn}. The VTM is mainly used to project multi-view camera features onto the BEV plane.

However, extreme weather, varied illumination, or noise confuse objects and backgrounds, affecting the network's perception capability. This phenomenon is particularly amplified in BEV representation, where features from all image regions are projected onto a single plane, causing severe information entanglement between foreground objects and complex backgrounds. Such spatial compression increases the difficulty of feature discrimination and leads to ambiguous positional priors for 3D detection. This motivates us to introduce 2D regions attention guidance to 1) enhance the backbone's capability in distinguishing critical objects through attention mechanisms, and 2) provide explicit structural priors from perspective-aware 2D regions results. Specifically, the per-view 2D regions establish reliable anchor points in image coordinates, which effectively constrain the search space for 3D detection and mitigate the learning ambiguity caused by BEV projection.

Therefore, in this work, we introduce a method called \textbf{2D Region-Oriented Attention for a BEV-based 3D Object Detection Network (ROA-BEV)}, intended to enable the image feature extractor of the network to focus more on learning where objects exist, thereby reducing interference from other background information. In order to generate more accurate regions, we directly use multi-scale features from the feature extractor and fuse the generated results. Meanwhile, each scale of network uses large kernel convolution to capture information. The convolutional kernel of the large receptive field better balances the learning background and foreground, as well as the contextual relationships between objects in the foreground. In summary, the major contributions of this paper are:
\begin{itemize}
    \item We propose ROA-BEV to enable the network to focus on extracting features of regions where objects exist and separate them from background,and it is applicable to the previous BEV methods. 
    \item We propose ROA to generate regions of objects in camera views, which fuses multi-scale features from the image backbone. Large kernel is used on every scale to catch more information, especially on large objects.
    \item Our method is validated to be effective in experiments based on BEVDepth \cite{bevdepth} on the nuScenes val set.
\end{itemize}


\section{Related Work}
\label{sec:formatting}
\subsection{Vision-based 3D Object Detection}
The key challenge of vision-based 3D object detection is the inherent ambiguity in estimating depth from images. Despite this, significant progress has been made through various approaches. One prominent direction involves predicting 3D bounding boxes from 2D image features. Early works, such as CenterNet \cite{centernet}, demonstrates that 2D detectors can be adapted for 3D detection with minimal modifications. FCOS3D \cite{fcos3d} converts 3D targets to the image domain for predicting both 2D and 3D attributes. DD3D \cite{dd3d} further emphasizes the benefit of depth pretraining.
\begin{figure*}[!ht]
    \centering
    \includegraphics[width=1\linewidth]{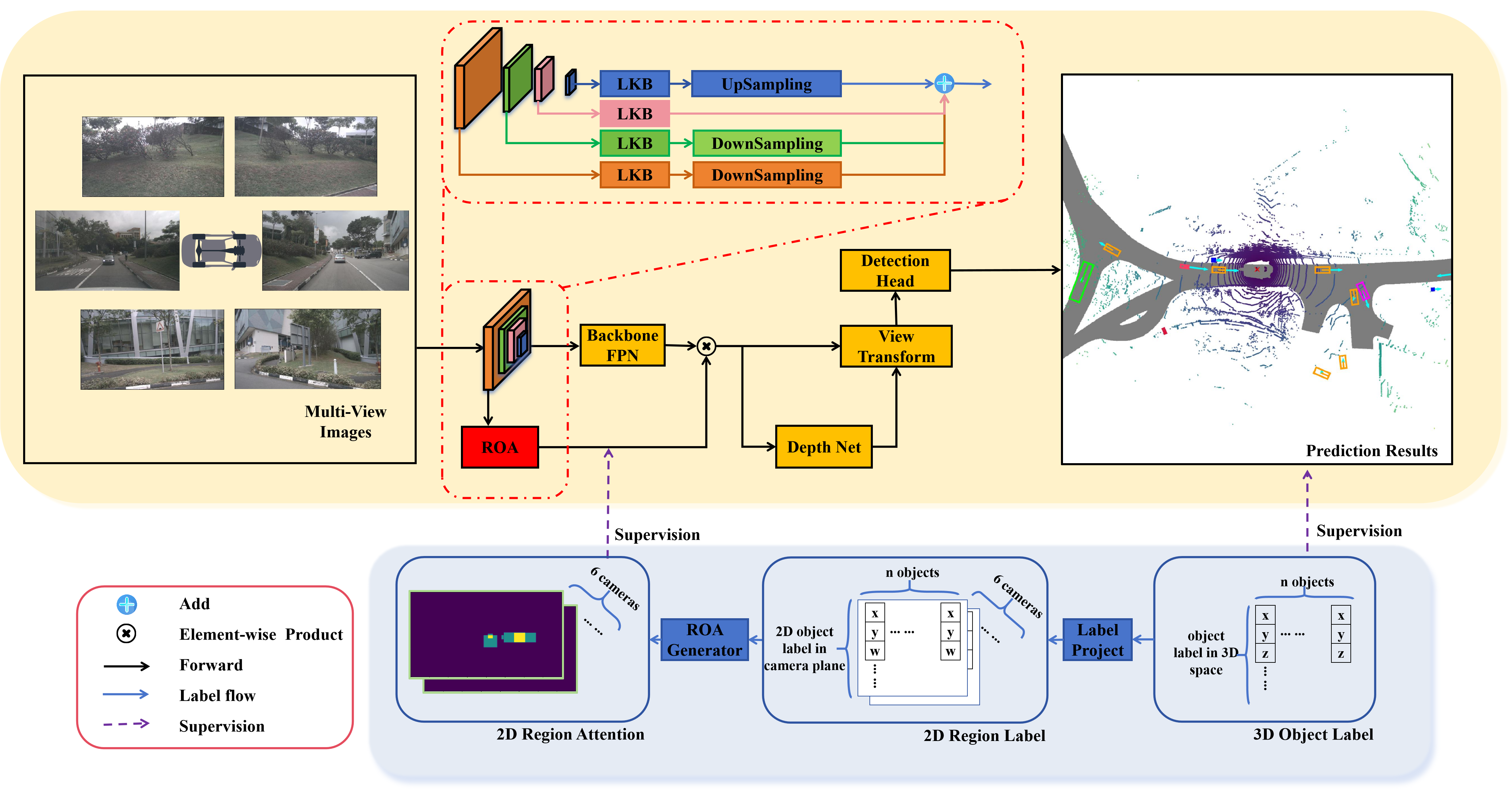}
    \caption{Overall framework of ROA-BEV. The ROA uses multi-scale image features to generate object regions from the view of cameras. The supervision of ROA is 2D region attention, which is projected and generated from the 3D object label.}
    \label{fig:overview}
\end{figure*}
Another active research line focuses on projecting 2D image features into 3D space. LSS pioneers the view transform method, predicting depth distributions and projecting features onto BEV. BEVDet \cite{bevdet} utilizes the BEV feature space for 3D detection. BEVDet4D \cite{bevdet4d} combines multiple frames of temporal information to perform joint spatiotemporal feature modeling in the BEV space, effectively improving the accuracy of dynamic obstacle detection and the ability to estimate motion states. DETR3D \cite{detr3d} and BEVFormer \cite{bevformer} employs object queries and deformable attention to generate BEV features, while BEVDepth \cite{bevdepth} applies explicit depth supervision to improve accuracy. PETR \cite{petr} further improves upon DETR3D \cite{detr3d} by incorporating 3D position-aware representations. However, the study of the image feature extractor also needs to be focused. This paper proposes ROA-BEV, which can be used in previous BEV-based methods to generate the region's attention.

\subsection{Large Kernel Network}
In the domain of computer vision, transformer-based models, such as Swin Transformer \cite{swin} has garnered considerable attention. Their success can be attributed to their extensive receptive fields, as evidenced by numerous studies. Recently, convolutional networks featuring carefully designs large receptive fields have emerged as formidable competitors to transformer-based models. For instance, ConvNeXt \cite{convnet} leverages 7×7 depth-wise convolutions, achieving notable performance enhancements in downstream tasks. In this paper, we utilize the large kernel to increase the receptive field to generate accurate regions for objects in camera views, especially for large objects, such as trucks.


\subsection{2D Auxiliary Tasks for 3D Detection}
DeepMANTA \cite{deepmanta} presents a coarse-to-fine architecture with 2D object labels as middle supervision. MonoPSR \cite{monopsr} utilizes detections from a mature 2D object detector to generate a 3D proposal per object in a scene through the fundamental relations of a pinhole camera model. GUPNet \cite{gupnet} uses ROIAlign to obtain ROI features from the results generated by the 2D detector, while the predictions of the 2D detector and the results of the 3D detector are gathered through Hierarchical Task Learning strategy to assign proper weights. Far3D \cite{far3d} generates reliable 2D box proposals and their corresponding depths, which are then concatenated and projected into 3D space. In this paper, 2D labels are produced as the region attention form to be applied on the image feature extractor. In addition, we consider the overlap between objects.

\section{Method}
\subsection{Overall Architecture}
Most networks employ classic feature extractors like ResNet \cite{resnet} as the backbone. However, the network's overall supervision is solely provided by the sparse labels in 3D object detection, preventing the feature extraction network from effectively focusing on areas with objects. We develop an ROA network to provide the network with prior knowledge of the object area from a 2D perspective. As shown in Fig. \ref{fig:overview}, the 2D ROA-BEV receives multi-view images as input.  The input is first processed through the backbone to extract features, followed by FPN that fuses features across various scales. To generate a region-oriented attention map, we design the ROA module. This module receives features from various scales rather than the fused features backbone FPN. The region-oriented map predicted by ROA is then multiplied by the image feature attention, along with the features from the FPN network, to produce features more focused on potential object areas. Subsequently, similar to BEVDepth, these image features are utilized for viewpoint transformation and subsequent detection head.
\subsection{Multi-scale 2D Region Oriented Attention}
In the forward feature extraction process, objects of different scales have different expressions in different layers, accompanied by information loss. We utilize features from four scales within the backbone, inputting each into the \textbf{Large Kernel Basic(LKB)} module. The detail of ROA is illustrated in the red dashed box of Fig. \ref{fig:overview}. Subsequently, four scaled features are either upsampled or downsampled to match the FPN network before being summed. The details of processing multi-scale 2D ROA can be formulated as:
\begin{equation}
\begin{split}
    \mathcal{O}_{ROA} &= DS(\mathcal{F}_{LKB}^{\xi_{1}}(\mathcal{O}_{backbone}^{1}),4) \\
    &+ DS(\mathcal{F}_{LKB}^{\xi_{2}}(\mathcal{O}_{backbone}^{2}),2) \\
    &+ \mathcal{F}_{LKB}^{\xi_{3}}(\mathcal{O}_{backbone}^{3}) \\
    &+ U(\mathcal{F}_{LKB}^{\xi_{4}}(\mathcal{O}_{backbone}^{4}),2)
\end{split}
\end{equation}
where $DS(input,s)$ denotes downsampling the features by $s$ times and $U(input,s)$ denotes upsampling the features by $s$ times. $\mathcal{F}_{LKB}^{\xi}$ represents the LKB module. $\mathcal{O}_{backbone}^{i}$ denotes the output of $i$-th scale in the backbone.

Finally, the generated 2D ROA will be element-wise multiplied by the output of FPN. The attention mechanism can be formulated as:
\begin{equation}
    \mathcal{O}_{feature} = \mathcal{O}_{FPN} \times \mathcal{O}_{ROA}
\end{equation}

\subsection{Large Kernel Basic Module}
\begin{figure}
    \centering
    \includegraphics[width=0.65\linewidth]{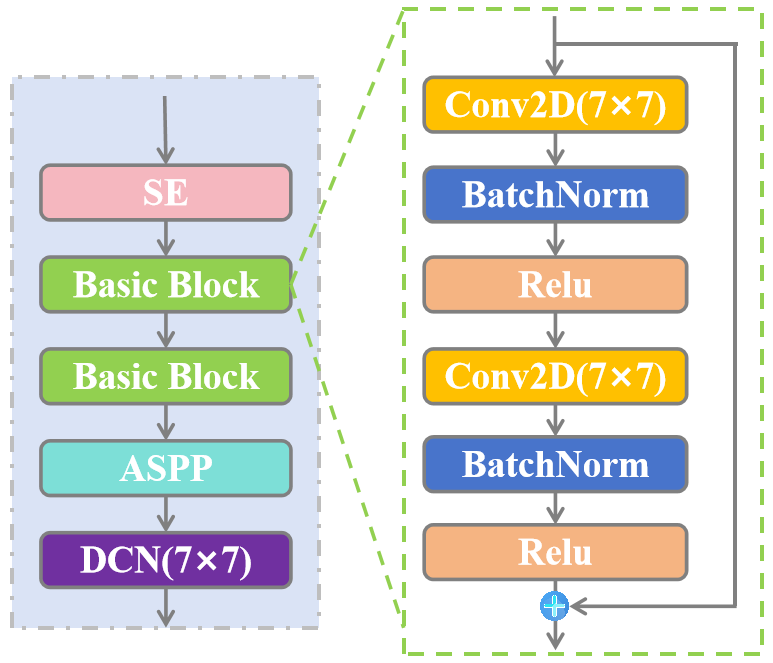}
    \caption{Details of the Large Kernel Basic (LKB) module in the ROA.}
    \label{fig:LKB}
\end{figure}
As shown in Fig. \ref{fig:LKB}, the Large Kernel Basic (LKB) module is modified on DepthNet in BEVDepth, which contains Squeeze-and-Excitation (SE) \cite{se}, Basic block, Atrous Spatial Pyramid Pooling (ASPP) \cite{aspp}, and Deformable Convolution Network (DCN) \cite{dcn}. However, different from the DepthNet, the kernel size of every basic block and DCN is $7 \times 7$. These large kernel convolutions, characterized by their extended receptive fields, enable the model to capture a broader context and spatial relationships within the input data, enhancing feature extraction capabilities. Although DCN introduces deformability to adapt to input shape changes, the benefits of the large kernel convolutions used earlier in the model are still evident, as they contribute to the overall robustness and feature discriminability. 


\subsection{2D ROA Label Generation}
\begin{algorithm}
\label{algorithm:2d label}
\caption{Ground Truth Generation of ROA}
\begin{algorithmic}[1]
\REQUIRE 3D labels of objects in one frame $List\{B_{3D}\}$, extrinsic $\{\mathcal{R}|\ \mathcal{T}\}$ and intrinsic $\{\mathcal{K}\}$ of cameras
\STATE Initialize $GT_{ROA}=0.1\times \mathbf{1}_{Cameras \times H \times W}$
\FOR{each 3D bounding box $B_{3D}$ in $List\{B_{3D}\}$}
    \FOR{each camera index $i$}
        \STATE Coordinate system transformation:
        \STATE $B' \leftarrow [\mathcal{R}_i|\ \mathcal{T}_i]B_{3D}$
        \STATE $B_{2D} \leftarrow \mathcal{K}_iB'$        
        \IF{$B_{2D}$ is within the $i$-th camera FOV}
            \STATE Calculate the corners of 2D bounding box in the camera $i$:
            \STATE $x_{\min}, y_{\min}, x_{\max}, y_{\max}  \leftarrow B_{2D}$

            \STATE Update GT of ROA:
            \STATE $GT_{ROA}[i][y_{\min}:y_{\max}, x_{\min}:x_{\max}] \mathrel{+}= 1.0$
            \ENDIF
    \ENDFOR
\ENDFOR
\RETURN The ground Truth of ROA $GT_{ROA}$
\end{algorithmic}
\end{algorithm}
As illustrated in the bottom blue box of Fig. \ref{fig:overview}, we transform the 3D coordinates of the target within the ego coordinate system into 2D coordinates in the camera coordinate system using the camera's extrinsic and intrinsic parameters. Subsequently, we initialize six matrices to all zeros, one for each camera, and draw frames in these matrices based on the 2D labels. It should be noted that in order to prevent the impact of all zero matrices on network convergence, we add 0.1 to the initialized matrix. The area within the frame is assigned the value of 1. If the same pixel is enclosed by different boxes multiple times, the values at that pixel are cumulatively added. The complete process is summarized in Algorithm 1.
\begin{table*}[!ht]
\centering
\caption{Comparison on the nuScenes val set. All results are trained with CBGS \cite{cbgs}}.
\begin{tabular}{c|c|c|cc|c|c|c|c|c}
\hline
\textbf{Method} & \textbf{Backbone} & \textbf{Resolution} & \textbf{mAP↑} & \textbf{NDS↑} & \textbf{mATE↓} & \textbf{mASE↓} & \textbf{mAOE↓} & \textbf{mAVE↓} & \textbf{mAAE↓} \\ \hline
\textbf{PETR \cite{petr}} & R50 & 384×1056 & 0.313 & 0.381 & 0.768 & 0.278 & 0.564 & 0.923 & 0.225 \\  
\textbf{BEVDet \cite{bevdet}} & R50 & 256×704 & 0.298 & 0.379 & 0.725 & 0.279 & 0.589 & 0.860 & 0.245 \\  
\textbf{BEVDet4D \cite{bevdet4d}} & R50 & 256×704 & 0.322 & 0.457 & 0.703 & 0.278 & 0.495 & 0.354 & 0.206 \\  
\textbf{BEVDepth \cite{bevdepth}} & R50 & 256×704 & 0.351 & 0.475 & \color[HTML]{fe0000}\textbf{0.639} & \color[HTML]{fe0000}\textbf{0.267} & 0.479 & 0.428 & \color[HTML]{fe0000}\textbf{0.198} \\ \hline
\textbf{ROA-BEV} & R50 & 256×704 & \color[HTML]{fe0000}\textbf{0.361} & \color[HTML]{fe0000}\textbf{0.485} & 0.640 & 0.269 & \color[HTML]{fe0000}\textbf{0.459} & \color[HTML]{fe0000}\textbf{0.374} & 0.212 \\ \hline
\end{tabular}
\label{table: val}
\end{table*}
\subsection{Training Loss}
This section introduces the loss function. As shown in Fig. \ref{fig:overview}, our method has three major branches to be supervised: Detection Head, DepthNet and ROA. Therefore, we define three functions to minimize: $\mathcal{L}_{Det}$, $\mathcal{L}_{Depth}$ and $\mathcal{L}_{ROA}$. The total loss can be formulated as:
\begin{equation}
    \mathcal{L}=\mathcal{L}_{Det}+\lambda_{1}\mathcal{L}_{Depth}+\lambda_{2}\mathcal{L}_{ROA}
\end{equation}
where $\lambda_{1}$ denotes the weight of depth loss, $\lambda_{2}$ denotes the weight of ROA loss.
$\mathcal{L}_{Det}$ and $\mathcal{L}_{Depth}$ follow BEVDepth.
$\mathcal{L}_{ROA}$ is the ROA loss function for the supervision of the proposed ROA branch. We use $\mathcal{L}$1 loss to minimize the $\mathcal{L}_{roa}$: 
\begin{equation}
    \mathcal{L}_{ROA}=\mathcal{L}_{1}\{\mathcal{O}_{ROA},GT_{ROA}\}
\end{equation}
where $GT_{ROA}$ denotes the ground truth of 2D Region Attention in the Fig. \ref{fig:overview}.

\subsection{Dataset and Metrics}
\section{Experiments}
\begin{figure*}
    \centering
    \includegraphics[width=0.85\linewidth]{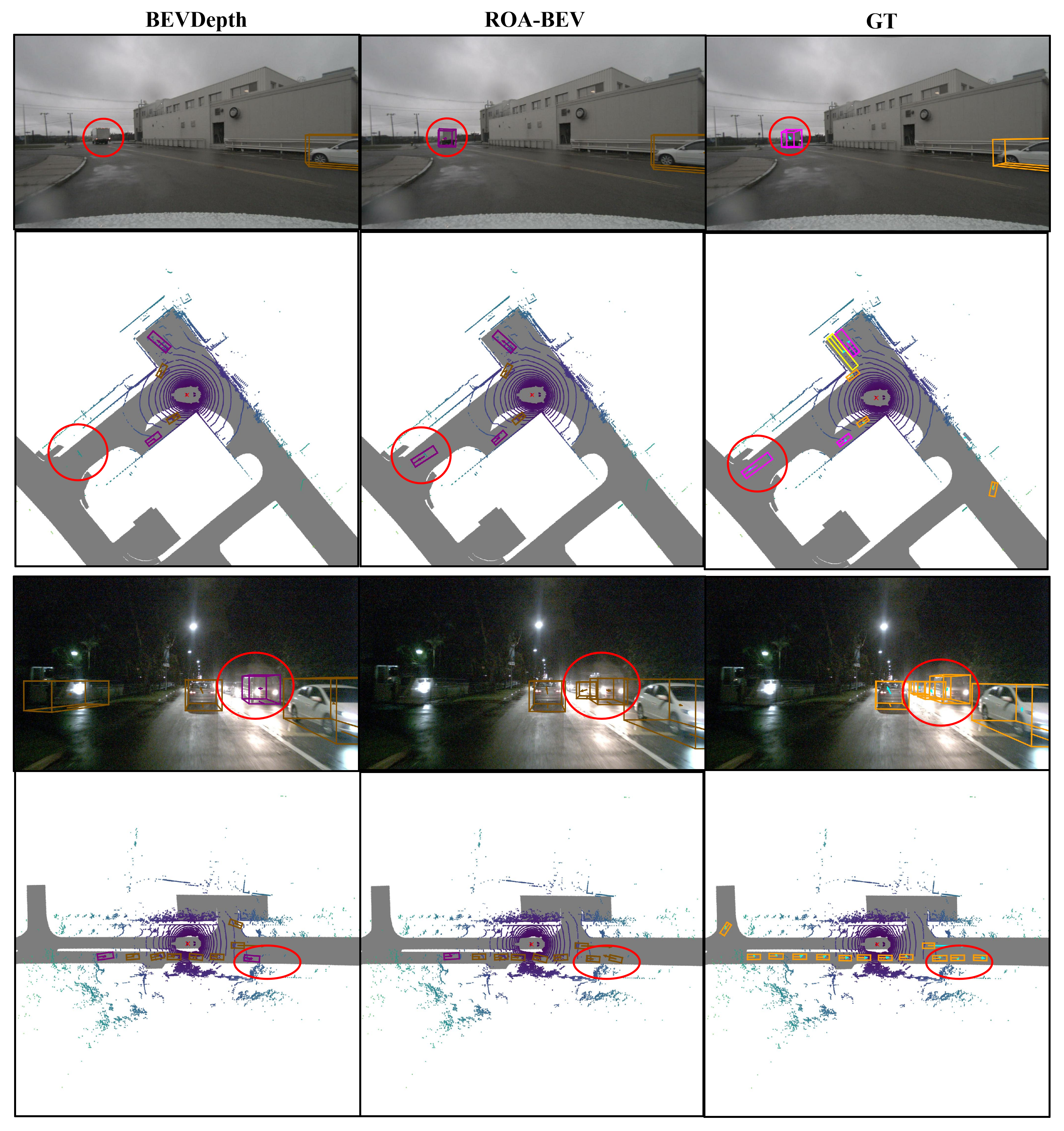}
    \caption{Visualization of detection results on images and BEV view.}
    \label{fig:results_compare_others}
\end{figure*}
The nuScenes dataset\footnote{https://www.nuscenes.org} \cite{nuscenes} stands as a significant benchmark for autonomous driving research. This dataset encompasses 750 scenarios for training, 150 for validation, and another 150 for testing. The data encompasses inputs from six cameras, one LiDAR, and five radars, capturing the multimodal nature of the driving environment.
For 3D object detection task, the nuScenes Detection Score (NDS) is a crucial metric, integrating various performance aspects beyond the traditional mean Average Precision (mAP). The NDS considers additional true positive metrics, including mean Average Translation Error (mATE), mean Average Scale Error (mASE), mean Average Orientation Error (mAOE), mean Average Velocity Error (mAVE), and mean Average Attribute Error (mAAE). 

\subsection{Implementation Details}
We utilize BEVDepth \cite{bevdepth} as our baseline. All experiments are trained and tested with 8 NVIDIA GeForce RTX 2080ti GPUs. Batch size is set to 2 in each GPU. We employ ResNet-50 \cite{resnet} as the image backbone and the image resolution is 704*256. Our models are trained using the AdamW \cite{adamw} optimizer, and gradient clipping is applied. When compared to other methods in the Table \ref{table: val}, all results are obtained with CBGS \cite{cbgs}. The epoch of the training stage is 20. The initial learning rate is 2e-4. Learning rate decays occur at epochs 10, 14, and 18. Each update reduces the learning rate to 0.6 times the previous rate. For the ablation study, all models are trained 40 epochs without CBGS, and the learning rate is 1.5e-4. Other settings are the same. $\lambda_{1}$ and $\lambda_{2}$ are set to 3 and 1 separately. All experiments are implemented based on MMDetecion3D \cite{mmdet3d}.

\subsection{Main Results}
\subsubsection{Comparison with State-of-the-Arts}
In our experiments, we apply CBGS \cite{cbgs} to various 3D object detection methods and evaluate their performance on a standard benchmark, which is presented in the Table \ref{table: val}. 
Our method achieves the highest mAP and NDS scores among all the compared methods, indicating its effectiveness in improving the overall detection performance. 



\subsubsection{Visualization}
\begin{figure}
    \centering
    \includegraphics[width=1\linewidth]{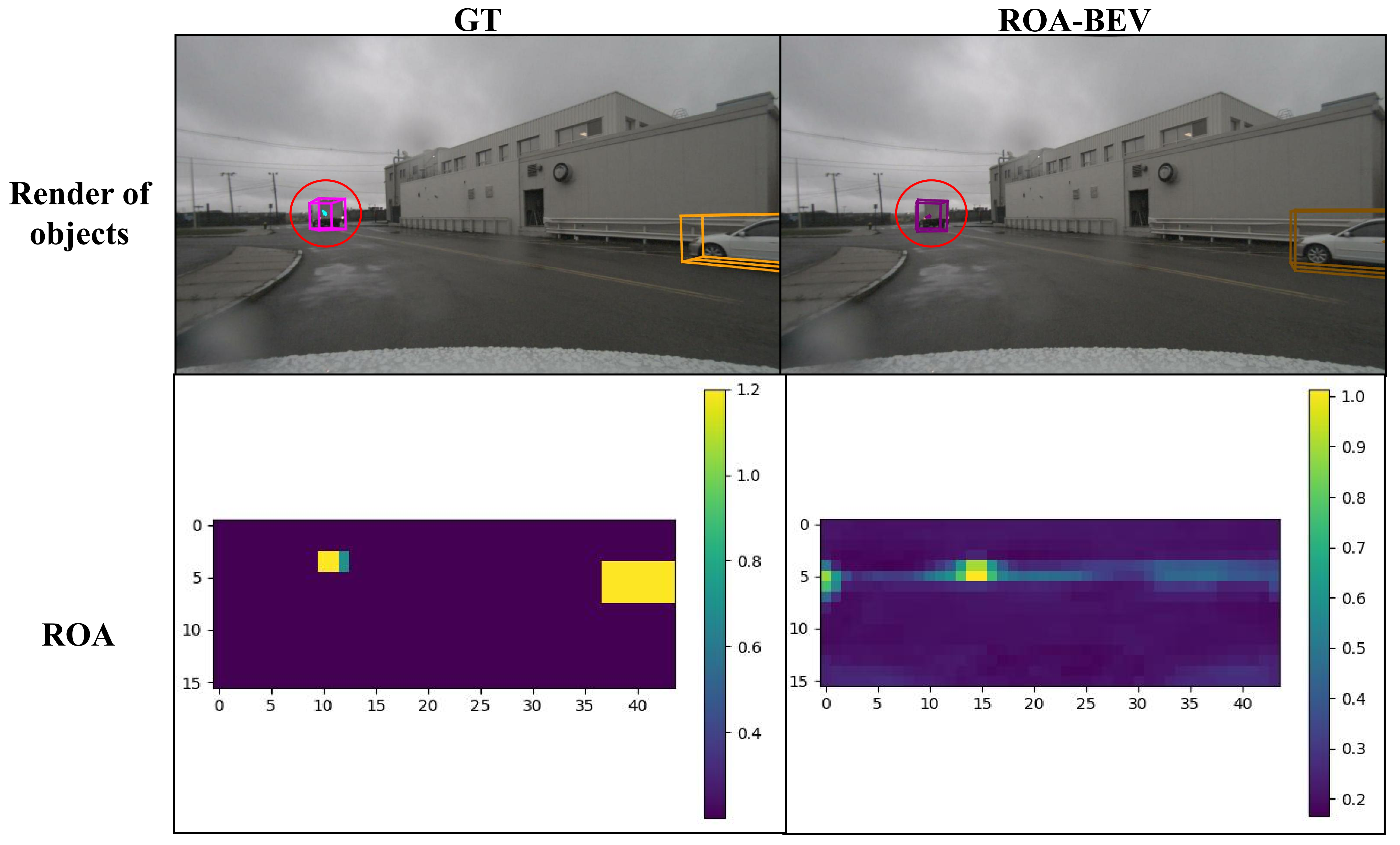}
    \caption{Visualization of ROA.}
    \label{fig:roa results}
\end{figure}

This section will show the visualization of detection results. The results are shown in the Fig. \ref{fig:results_compare_others}. We provide two examples to demonstrate the effectiveness of our method. In each example, the advantages of our method are marked with red circles. In the first example, the truck in the distance merges with the white clouds in the sky and the black ground. Our method successfully detects the truck. Similarly, our method successfully detects objects under backlighting in the second example. Moreover, the category is correct for cars, while BEVDepth's prediction results miss the car and identify another car as a truck.
To verify the effectiveness of ROA, we visualize the ROA results generated by the network. As shown in the Fig. \ref{fig:roa results}, the distant truck region can be generated. The distant truck blends well with the background, resulting in poor feature extraction from the backbone. Therefore, the output of ROA increases the weight of the corresponding position accordingly. On the contrary, the backbone can be extracted better for the car on the right, and the ROA results give smaller weights to compensate for the features.

\subsection{Ablation Study}

\subsubsection{Component Analysis}

\begin{table}[!ht]
\centering
\caption{Ablation study of using ROA and type of region on nuScenes \textit{val}.}
\begin{tabular}{c|cc|cc}
\hline
{\color[HTML]{000000} \textbf{}} & {\color[HTML]{000000} \textbf{ROA}} & {\color[HTML]{000000} \textbf{Type of Region}} & {\color[HTML]{000000} \textbf{mAP↑}} & {\color[HTML]{000000} \textbf{NDS↑}} \\ \hline
{\color[HTML]{000000} \textbf{Baseline}} & {\color[HTML]{000000} ×} & {\color[HTML]{000000} ×} & {\color[HTML]{000000} 0.329} & {\color[HTML]{000000} 0.443}  \\ \hline
{\multirow{5}*{\color[HTML]{000000} \textbf{Ours}}} & {\color[HTML]{000000} \checkmark} & {\color[HTML]{000000} ×} & {\color[HTML]{000000} 0.335} & {\color[HTML]{000000} 0.450}   \\ 
 & {\color[HTML]{000000} \checkmark} & {\color[HTML]{000000} Binary-Pre} & {\color[HTML]{000000} 0.338} & {\color[HTML]{000000} 0.454}   \\ 
 & {\color[HTML]{000000} \checkmark} & {\color[HTML]{000000} Overlap-Pre} & {\color[HTML]{000000} 0.349} & {\color[HTML]{000000} 0.461} \\ \cline{2-5}
 & {\color[HTML]{000000} ×} & {\color[HTML]{000000} Binary-GT} & {\color[HTML]{000000} 0.374} & {\color[HTML]{000000} 0.471}   \\ 
 & {\color[HTML]{000000} ×} & {\color[HTML]{000000} Overlap-GT} & {\color[HTML]{fe0000} \textbf{0.411}} & {\color[HTML]{fe0000} \textbf{0.490}}  \\ \hline
\end{tabular}
\label{table: ablation roa}
\end{table}
We evaluate the performance of our proposed method against the baseline BEVDepth model on various metrics. The results in the Table \ref{table: ablation roa} demonstrate that incorporating attention supervision significantly enhances the model's performance. Specifically, using the proposed ROA module but without additional ground truth (GT) supervision achieves an improvement in mAP (from 0.329 to 0.335) and NDS (from 0.443 to 0.450) compared to the baseline, BEVDepth, indicating better overall detection accuracy. Furthermore, when using the binary ROA lable to supervise the ROA, mAP increases to 0.338 and NDS to 0.454. Detailly, the binary label means changing the value of the overlap label to binarization. The above results show that using attention supervision can improve the detection results of the network. Specifically, if multiple objects overlap, using overlap instead of binary GT supervision can avoid the network mistaking multiple objects for one and improve network performance.

Futhermore, we experiment with the upper limit of network performance using ROA, which is the GT directly inputted into ROA. The trend of results is the same: with the use of overlap, it can achieve optimal network performance. In addition, by comparing the results predicted using GT and network, it can be seen that there is still a gap of 0.062 and 0.029 in mAP and NDS. This indicates that the method has the potential for further research and development.

\subsubsection{Specific Classes Analysis}
\begin{table*}[!h]
\centering
\caption{Ablation study of using ROA on different methods. "construc." denotes the category of construction vehicle.}
\begin{tabular}{c|c|cccc|cccc}
\hline
 & \textbf{} & \textbf{BEVDet \cite{bevdet}} & \textbf{\begin{tabular}[c]{@{}c@{}}BEVDet \cite{bevdet}\\ +ROA\end{tabular}} & \textbf{Difference} & \textbf{Percentage} & \textbf{BEVDepth \cite{bevdepth}} & \textbf{\begin{tabular}[c]{@{}c@{}}BEVDepth \cite{bevdepth}\\ +ROA\end{tabular}} & \textbf{Difference} & \textbf{Percentage} \\ \hline
\multirow{11}*{\textbf{mAP}} & \textbf{car} & 0.517 & {\color[HTML]{FE0000} \textbf{0.528}} & 0.011 & 2.13\% & 0.493 & {\color[HTML]{FE0000} \textbf{0.511}} & 0.018 & 3.74\% \\
 & \textbf{truck} & 0.226 & {\color[HTML]{FE0000} \textbf{0.243}} & 0.017 & 7.52\% & 0.258 & {\color[HTML]{FE0000} \textbf{0.280}} & 0.0221 & 8.57\% \\
 & \textbf{bus} & 0.305 & {\color[HTML]{FE0000} \textbf{0.328}} & 0.023 & 7.54\% & 0.362 & {\color[HTML]{FE0000} \textbf{0.399}} & 0.037 & 10.28\% \\
 & \textbf{trailer} & 0.101 & {\color[HTML]{FE0000} \textbf{0.105}} & 0.004 & 3.96\% & 0.153 & {\color[HTML]{FE0000} \textbf{0.18}} & 0.027 & 17.57\% \\
 & \textbf{construc.} & 0.039 & {\color[HTML]{FE0000} \textbf{0.044}} & 0.005 & 12.82\% & 0.056 & {\color[HTML]{FE0000} \textbf{0.068}} & 0.012 & 21.86\% \\
 & \textbf{pedestrian} & 0.318 & {\color[HTML]{FE0000} \textbf{0.340}} & 0.022 & 6.92\% & 0.357 & {\color[HTML]{FE0000} \textbf{0.375}} & 0.018 & 4.92\% \\
 & \textbf{motorcycle} & 0.216 & {\color[HTML]{FE0000} \textbf{0.241}} & 0.025 & 11.57\% & 0.308 & {\color[HTML]{FE0000} \textbf{0.316}} & 0.008 & 2.46\% \\
 & \textbf{bicycle} & {\color[HTML]{FE0000} \textbf{0.203}} & 0.199 & -0.004 & -1.97\% & 0.308 & {\color[HTML]{FE0000} \textbf{0.311}} & 0.003 & 1.04\% \\
 & \textbf{traffic cone} & 0.499 & {\color[HTML]{FE0000} \textbf{0.511}} & 0.012 & 2.40\% & 0.492 & {\color[HTML]{FE0000} \textbf{0.524}} & 0.032 & 6.57\% \\
 & \textbf{barrier} & 0.404 & {\color[HTML]{FE0000} \textbf{0.436}} & 0.0320 & 7.92\% & 0.505 & {\color[HTML]{FE0000} \textbf{0.523}} & 0.019 & 3.67\% \\
 & \textbf{Average} & 0.283 & {\color[HTML]{FE0000} \textbf{0.298}} & 0.015 & 5.20\% & 0.329 & {\color[HTML]{FE0000} \textbf{0.349}} & 0.020 & 5.92\% \\ \hline
\textbf{NDS} & \textbf{Average} & 0.350 & {\color[HTML]{FE0000} \textbf{0.368}} & 0.018 & 5.11\% & 0.443 & {\color[HTML]{FE0000} \textbf{0.461}} & 0.018 & 3.95\% \\ \hline
\end{tabular}
\label{table:ablation roa on dif methods}
\end{table*}

In our experiments, we evaluate the performance of our proposed method by integrating it with two 3D object detection frameworks: BEVDet \cite{bevdet} and BEVDepth \cite{bevdepth}. The results, presented in the TABLE \ref{table:ablation roa on dif methods}, demonstrate the effectiveness of our approach in enhancing detection accuracy across various categories of objects. For BEVDet, our method consistently improves the detection accuracy for most object categories, with notable percentage increases in categories such as 'construction\_vehicle' (+12.82\%) and 'motorcycle' (+11.57\%). Similarly, when integrated with BEVDepth, our approach leads to substantial improvements, particularly in categories like 'construction\_vehicle' (+21.86\%), 'trailer' (+17.57\%) and 'bus' (+10.28\%). It is worth noting that while there is a minor decrease in performance for one category, 'bicycle' with BEVDet. We consider that the possible reason is that their thin, narrow shape gets more easily distorted when converting to the BEV, combined with fewer bicycle examples in our training data. 
However, the overall trend of the mAP and NDS scores indicates a positive impact on detection accuracy, with improvements of 5.20\% and 5.11\% respectively for BEVDet, and 5.92\% and 3.95\% for BEVDepth. These results underscore the efficacy of our method in enhancing the performance of existing 3D object detection models, particularly in challenging scenarios involving small or partially occluded objects.

\subsubsection{Multi-scale Analysis}
We analyze the impact of different feature extraction methods. The results in the Table \ref{table:scale} indicate that incorporating multi-scale features yields the best performance. These findings underscore the importance of utilizing multi-scale features directly from the backbone, as this structure can reduce the loss of information transmission after FPN.
\begin{table}[!ht]
\centering
\caption{Ablation study of using multi-scale features.}
\begin{tabular}{c|c|c}
\hline
\textbf{Input of the ROA feature} & \textbf{mAP↑} & \textbf{NDS↑} \\ \hline
\textbf{Same scale from backbone} & 0.332 & 0.451 \\ \hline
\textbf{FPN backbone} & 0.345 & 0.453 \\ \hline
\textbf{Multi-scale from backbone} & {\color[HTML]{FE0000} \textbf{0.349}} & {\color[HTML]{FE0000} \textbf{0.461}} \\ \hline
\end{tabular}
\label{table:scale}
\end{table}

\subsubsection{Kernel Size Analysis}

\begin{table}[!ht]
\centering
\caption{Ablation study of using different size of kernel.}
\begin{tabular}{c|cc}
\hline
\textbf{Kernel size of basic block} & {\color[HTML]{000000} \textbf{mAP↑}} & \textbf{NDS↑} \\ \hline
\textbf{3*3} & 0.342 & 0.457 \\ \hline
\textbf{5*5} & 0.339 & 0.450 \\ \hline
\textbf{7*7} & {\color[HTML]{FE0000} \textbf{0.349}} & {\color[HTML]{FE0000} \textbf{0.461}} \\ \hline
\textbf{9*9} & 0.335 & 0.450 \\ \hline
\textbf{11*11} & 0.334 & 0.449 \\ \hline
\end{tabular}
\label{table:kernel size}
\end{table}

We investigate the impact of varying kernel sizes in the basic block. The results, presented in the Table \ref{table:kernel size}, a kernel size of 7 achieves the highest mAP and NDS with a value of 0.349 and 0.461 separately. This suggests that the $7 \times 7$ kernel size aids in capturing more contextual information, simultaneously balancing parameter quantity and accuracy

\subsubsection{Results Distribution Ayalysis}
\begin{figure}
    \centering
    \includegraphics[width=1\linewidth]{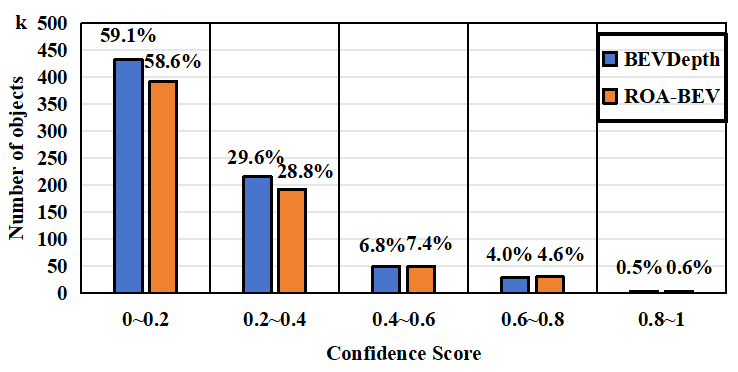}
    \caption{Comparison of the distribution of results. The total results predicted by BEVDepth \cite{bevdepth} and ROA-BEV are 731992 and 667578, respectively.}
    \label{fig:distribution}
\end{figure}

As shown in the Fig. \ref{fig:distribution}, the number of predicted results has significantly decreased. Meanwhile, the proportion of high confidence prediction results has also increased. 

\subsection{Conclusion}
In this paper, we propose ROA-BEV to orient feature extraction to focus on object regions in camera views. ROA fusions multi-scale features of images to generate attention regions. LKB in ROA increases the receptive field to enhance the network's performance between the background and objects, as well as the relationships between different objects. The ROA supervisor does not need additional data because the ROA label is projected from a 3D label. ROA-BEV can be embedded into most BEV-based 3D object detection methods.

However, some limitations also exist in our method. The performance of ROA has a significant impact on the final 3D object detection results. Errors and omissions in the region can reinforce unfavorable information and mislead the network. Given that there is still a significant gap between the existing results and the use of GT, improving network efficiency is an area that can be studied in the future.

\newpage
\bibliographystyle{IEEEtran}
\bibliography{IEEEfull}

\end{document}